\documentclass[letterpaper]{article} 
\usepackage{aaai23}  
\usepackage{times}  
\usepackage{helvet}  
\usepackage{courier}  
\usepackage[hyphens]{url}  
\usepackage{graphicx} 
\usepackage{natbib}  
\usepackage{caption} 
\usepackage[dvipsnames]{xcolor}
\usepackage{multirow}
\usepackage{makecell}
\usepackage{colortbl}
\usepackage{amsfonts}
\usepackage{amsmath}
\usepackage{amsthm}
\usepackage{algorithm2e}
\RestyleAlgo{ruled}
\usepackage{bbm}

\newcommand\R{\mathbb{R}}

\usepackage{xspace}
  \newcommand{\eg}{e.\,g.\xspace}
  \newcommand{\ie}{i.\,e.\xspace}
  
\newcommand{\uproman}[1]{\uppercase\expandafter{\romannumeral#1}}
\newcommand{\lowroman}[1]{\romannumeral#1\relax}
  
\newcommand\E{\mathrm{E}}

\newcommand\Indicator{\mathbbm{1}}
\newcommand{\indep}{\perp \!\!\! \perp}

\usepackage{booktabs}

\title{Estimating Average Causal Effects from Patient Trajectories}
\author {
    Dennis Frauen,\textsuperscript{\rm 1, 2}
    Tobias Hatt, \textsuperscript{\rm 3}
    Valentyn Melnychuk \textsuperscript{\rm 1, 2}
    Stefan Feuerriegel \textsuperscript{\rm 1, 2}
}
\affiliations {
    \textsuperscript{\rm 1} LMU Munich\\
    \textsuperscript{\rm 2} Munich Center for Machine Learning\\
    \textsuperscript{\rm 3} ETH Zurich\\
    frauen@lmu.de, thatt@ethz.ch, melnychuk@lmu.de, feuerriegel@lmu.de
}

\usepackage{bibentry}

\begin{document}

\maketitle

\begin{abstract}
In medical practice, treatments are selected based on the expected causal effects on patient outcomes. Here, the gold standard for estimating causal effects are randomized controlled trials; however, such trials are costly and sometimes even unethical. Instead, medical practice is increasingly interested in estimating causal effects among patient (sub)groups from electronic health records, that is, observational data. In this paper, we aim at estimating the average causal effect (ACE) from observational data (patient trajectories) that are collected over time. For this, we propose DeepACE: an end-to-end deep learning model. DeepACE leverages the iterative G-computation formula to adjust for the bias induced by time-varying confounders. Moreover, we develop a novel sequential targeting procedure which ensures that DeepACE has favorable theoretical properties, \ie, is doubly robust and asymptotically efficient. To the best of our knowledge, this is the first work that proposes an end-to-end deep learning model tailored for estimating time-varying ACEs. We compare DeepACE in an extensive number of experiments, confirming that it achieves state-of-the-art performance. We further provide a case study for patients suffering from low back pain to demonstrate that DeepACE generates important and meaningful findings for clinical practice. Our work enables practitioners to develop effective treatment recommendations based on population effects.
\end{abstract}

\section{Introduction}

Causal effects of treatments on patient outcomes are hugely important for decision-making in medical practice \cite{Yazdani.2015, Melnychuk.2022b}. These estimates inform medical practitioners in the expected effectiveness of treatments and thus guide treatment selection. Notwithstanding, information on causal effects is also relevant for other decision-making of other domains such as public health \cite{Glass.2013} and marketing \cite{Varian.2016}. 

Causal effects can be estimated from either randomized controlled trials (RCTs) or observational studies \cite{Robins.2000, Frauen.2022}. Even though RCTs present the gold standard for estimating causal effects, they are often highly costly, unfeasible in practice, or even unethical (\eg, medical professionals cannot withhold effective treatment to patients in need) \cite{Robins.2000}. Therefore, medical practice is increasingly relying on observational data to study causal relationships between treatments and patient outcomes. Nowadays, electronic health records are readily available, capturing patient trajectories with high granularity, and thus providing rich observational data in medical practice \cite{Allam.2021}.


In this paper, we aim at estimating causal effects from observational data in form of patient trajectories. Patient trajectories encode medical histories in a time-resolved manner and are thus of longitudinal form. However, estimating causal effects from observational data is subject to challenges \cite{Robins.2009}. The reason is that the underlying treatment assignment mechanism is usually confounded with the patient outcomes. Another reason is that  confounders may vary over time, which introduces additional dependencies and treatment-confounder feedback. 


While there are works on estimating individualized causal effects \cite{Lim.2018, Bica.2020, Melnychuk.2022}, we are interested in \textbf{{average} causal effects (ACEs)}. As such, ACEs give the expected difference in health outcomes when applying different treatment interventions at the population level. ACEs are important in many applications ranging from marketing to epidemiology, where policies are always based on (sub)population effects \cite{Varian.2016, Naimi.2017}. A prominent example is public health: here, a government might be interested in the different average effects of a stay-home order on COVID-19 spread for vaccinated vs. non-vaccinated people.


\textbf{Proposed method}: We propose an end-to-end deep learning model to estimate time-varying ACEs, called DeepACE. DeepACE combines a recurrent neural network and feed-forward neural networks to learn conditional expectations of factual and counterfactual outcomes under complex non-linear dependencies, based on which we then estimate time-varying ACEs. In DeepACE, we address time-varying confounding by leveraging the G-formula, which expresses the ACE as a sequence of nested conditional expectations based on observational data as. Existing methods are limited in that these learn the nested conditional expectations \emph{separately} by performing an iterative procedure \cite{vanderLaan.2018}. In contrast, our end-to-end model DeepACE makes it possible to learn them \emph{jointly}, leading to a more efficient use of information across time.  


We further develop a \emph{sequential targeting procedure} by leveraging results from semi-parametric estimation theory in order to improve the estimation quality of DeepACE. The sequential targeting procedure perturbs (``{targets}'') the outputs of DeepACE so that our estimator satisfies a semi-parametric efficient estimating equation. To achieve this, we propose a targeting layer and a targeted regularization loss for training. We then derive that DeepACE provides a doubly robust and asymptotically efficient estimator.


Our main \textbf{contributions}\footnote{Code available at \url{https://github.com/DennisFrauen/DeepACE}.}: (i)~We propose DeepACE: the first end-to-end neural network for estimating time-varying average causal effects using observational data. DeepACE builds upon the iterative G-computation formula to address time-varying confounding. (ii)~We develop a novel sequential targeting procedure which ensures that DeepACE provides a doubly robust and asymptotically efficient estimator. (iii)~We perform an extensive series of computational experiments using state-of-the-art models for time-varying ACE estimation, establishing that DeepACE achieves a superior performance. We further demonstrate that DeepACE generates important findings based on a medical case study for patients suffering from low back pain.

\section{Related work}
\label{sec:related_work}

Estimating causal effects from observational data can be grouped into static and longitudinal settings (Table~\ref{t:rel_work}).

\begin{table}[htbp]
    \vspace{-0.2cm}
    \caption{Key methods for causal effect estimation. \colorbox{OliveGreen!20}{This paper:} {ACE} for {longitudinal settings}.}
    \label{t:rel_work}
    \vspace{-0.2cm}
    \centering
    \resizebox{\columnwidth}{!}{\begin{tabular}{rl|l|l}
        \noalign{\smallskip} \toprule \noalign{\smallskip}
        & & \textbf{Static setting} & \textbf{{Longitudinal setting}}\\
        \midrule
        \multirow{4}{*}{\rotatebox[origin=c]{90}{\textbf{Causal effects}}} &
        \multirow{2}{*}{\makecell{Individual \\ (=ITE)}} & \eg, TARNet & RMSNs \cite{Lim.2018},\\
        & & \cite{Shalit.2017}, &  CRN \cite{Bica.2020}, \\
        & & causal forest\cite{Wager.2018} & G-Net \cite{Li.2021}\\
        \cmidrule{2-4}
        & \multirow{2}{*}{\makecell{{Average} \\ {(=ACE)}}}& \eg, TMLE \cite{vanderLaan.2006}, & \cellcolor{OliveGreen!20} g-methods \cite{Naimi.2017}, \\
        & & DragonNet \cite{Shi.2019} & \cellcolor{OliveGreen!20} LTMLE \cite{vanderLaan.2012},\\
        & &  & \cellcolor{OliveGreen!20} \textbf{DeepACE} (ours)\\
        \bottomrule
    \end{tabular}}
    \vspace{-0.4cm}
\end{table}

\subsection{Causal effect estimation in the static setting}

Extensive works focus on treatment effect estimation in static settings. Two important methods that adopt machine learning for \emph{average} treatment effect estimation in the static setting are: (\lowroman{1}) targeted maximum likelihood estimation (TMLE) \cite{vanderLaan.2006} and (\lowroman{2}) DragonNet \cite{Shi.2019}. TMLE is a plugin estimator that takes a (machine learning) model as input and perturbs the predicted outcomes, so that the final estimator satisfies a certain efficient estimating equation. This idea builds upon semi-parametric efficiency theory and is often called \emph{targeting}. Any estimator satisfying the efficient estimating equation is guaranteed to have desirable asymptotic properties. On the other hand, DragonNet is a neural network that incorporates a targeting procedure into the training process by extending the network architecture and adding a tailored regularization term to the loss function. This allows the model parameters to adapt simultaneously to provide a targeted estimate. To the best of our knowledge, there exists no similar targeting procedure for longitudinal data, and, to fill this gap, we later develop a tailored \emph{sequential} targeting procedure.

In general, causal effect estimation for static settings is different from longitudinal settings due to time-varying confounding and treatment-confounder feedback \cite{Robins.2009}.
Hence, methods for static causal effect estimation are biased when they are applied to longitudinal settings, thus leading to an inferior performance. For this reason, we later use methods for time-varying causal effect estimation as our prime baselines. Results for static baselines are reported in the Appendix\footnote{Appendix available at \url{https://arxiv.org/abs/2203.01228}.}.

\subsection{Causal effect estimation in longitudinal settings}

Causal effect estimation in longitudinal settings is often also called \emph{time-varying} causal effect estimation. Here, we distinguish individual and average causal effects.

\textbf{Individual causal effects:} There is a growing body of work on adapting neural networks for estimating \emph{individual} causal effects in the longitudinal setting (also individual treatment effects or ITE). These methods predict counterfactual outcomes conditioned on individual patient histories. Recurrent marginal structural networks (RMSNs) \cite{Lim.2018} use inverse probability weighting to learn a sequence-to-sequence model that addresses bias induced by time-varying confounders. The counterfactual recurrent network (CRN) \cite{Bica.2020} adopts adversarial learning to build balanced representations. The G-Net \cite{Li.2021} incorporates the G-formula into a recurrent neural network and applies Monte Carlo sampling. 

\emph{ITE methods are not optimal for ACE estimation}: Even though ITE methods can be used for ACE estimation by averaging individual effects, this is \textbf{not} optimal for two reasons: (i)~There is a well-established efficiency theory for ACE estimation \cite{Kennedy.2016}. In particular, methods that average ITE estimators suffer from so-called \emph{plug-in bias} \cite{Curth.2020}. (ii)~ITE methods for the longitudinal setting exclude covariates at time steps after the start of intervention from training and prediction (often by using an encoder-decoder architecture) in order to estimate effects of different interventions on individual patients. This is because these methods aim at estimating individual effects for unseen patients or time steps (out-of-sample). G-computation for unseen patients is hard because post-intervention covariates need to be predicted, leading to problems if X is high-dimensional. This is not needed for the ACE as we average over the observed population. Still, we later include the above state-of-the-art methods from ITE estimation (\ie, RMSNs, CRN, and G-Net) as baselines (we average individual estimates). There are other methods for predicting counterfactual outcomes over time (\eg, \cite{Schulam.2017, Soleimani.2017, Qian.2021, Berrevoets.2021}), which are not applicable due to different settings or assumptions.

\textbf{Average causal effects:} Several methods for time-varying ACE estimation originate from epidemiological literature. Here, common are so-called g-methods \cite{Naimi.2017}. Examples of g-methods include marginal structural models \cite{Robins.2000}, G-computation via the G-formula \cite{Robins.2009}, and structural nested models \cite{Robins.1994}. The previous methods make linearity assumptions and are consequently not able to exploit nonlinear dependencies within the data. Nevertheless, we include the g-methods as baselines.

We are aware of only one work that leverages machine learning methods for time-varying ACE estimation: longitudinal targeted maximum likelihood estimation (LTMLE) \cite{vanderLaan.2012}. LTMLE has two components: (i)~LTMLE uses iterative G-computation. Mathematically, the G-formula can be expressed in terms of nested conditional expectations, which can then be learned successively by using arbitrary regression models. As such, each conditional expectation is learned \emph{separately}, which is known as iterative G-computation. Our method, DeepACE, follows a similar approach but estimates the conditional expectations \emph{jointly}. (ii)~LTMLE targets the estimator by perturbing the predicted outcomes in each iteration to make them satisfy an efficient estimating equation. In contrast to LTMLE, our sequential targeting procedure is incorporated into the model training process, which allows all model parameters to be learned simultaneously. Later, we implement two variants of LTMLE as baselines, namely LTMLE with a generalized linear model (glm) \cite{vanderLaan.2012} and LTMLE with a super learner \cite{vanderLaan.2018}. 

\textbf{Research gap:} To the best of our knowledge, there exists no \emph{end-to-end} machine learning model tailored for \emph{time-varying ACE estimation}. Hence, DeepACE is the first neural network that simultaneously learns to perform iterative G-computation and to apply a sequential targeting procedure. By learning all parameters jointly, we expect our end-to-end model to provide more accurate estimation results.

\section{Problem Setup}
\label{sec:prob_setup}

\subsection{Setting}
\label{sec:setting}


We build upon the standard setting for estimating time-varying ACEs \cite{Robins.2009, vanderLaan.2012}. For each time step $t \in \{1,\dots,T\}$, we observe (time-varying) patient covariates $X_t \in \R^p$, treatments $A_t \in \{0,1\}$, and outcomes $Y_{t+1} \in \R$. For example, we would model critical care for COVID-19 patients by taking blood pressure and heart rate as time-varying patient covariates, ventilation as treatment, and respiratory frequency as outcome. Modeling the treatments $A_t$ as binary variables is consistent with prior works \cite{vanderLaan.2006, Shi.2019} and is standard in medical practice \cite{Robins.2000}: should one apply a treatment or not? 

\begin{figure}
\centering
\includegraphics[width=0.6\linewidth]{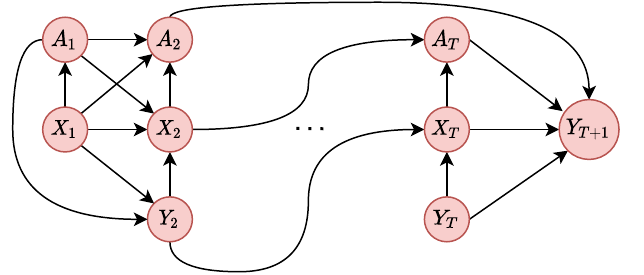}
\caption{One possible example of a causal graph describing the data-generating process.}
\label{fig:causal_graph}
\vspace{-0.4cm}
\end{figure}

At each time step $t$, the treatment $A_t$ directly affects the next outcome $Y_{t+1}$, and the covariates $X_t$ may affect both $A_t$ and $Y_{t+1}$. All $X_t$, $A_t$, and $Y_{t+1}$ may have direct effects on future treatments, covariates, and outcomes. The corresponding causal graph is shown in Fig.~\ref{fig:causal_graph}. For notation, we denote the observed trajectory at time $t$ by $\mathcal{H}_t = (\bar{X}_t,\bar{A}_{t-1})$, where $\bar{X}_t = (X_1,\dots,X_t)$ and $\bar{A}_{t-1} = (A_1,\dots,A_{t-1})$. We always consider the lagged outcomes $Y_t$ to be included in the covariates $X_t$.

We have further access to an observational dataset $\mathcal{D}$, that consists of $N$ independent patient trajectories $\mathcal{D}$ for patients $i \in \{1,\dots,N\}$, \ie, $\mathcal{D} = (\{x_t^{(i)}, a_t^{(i)}, y_{t+1}^{(i)}\}_{t=1}^T)_{i=1}^N$. Such patient trajectories are nowadays widely available in electronic health records \cite{Allam.2021}. For notation, we use a superscript $(i)$ to refer to patients (we omit it unless needed).


We build upon the potential outcomes framework \cite{Rubin.1978} and denote $Y_{t+1}\left(\bar{a}_{t} \right)$ as the potential outcome, which would have been observed at time $t+1$ if a treatment intervention $\bar{a}_t = (a_1,\dots,a_{t})$ was applied. Note that $Y_{t+1}\left(\bar{a}_{t} \right)$ is unobserved if the treatment intervention $\bar{a}_{t}$ does not coincide with the treatment assignments $\bar{A}_t$ in the observational dataset. This is also known as the fundamental problem of causal inference \cite{Pearl.2009b}.

\subsection{ACE estimation}

Given two intended treatment interventions $\bar{a}_{T} = (a_1,\dots,a_{T})$ and $\bar{b}_{T} = (b_1,\dots,b_{T})$, we define the expected potential outcomes as
\begin{equation}
    \theta^a = \E\left[Y_{T+1}\left(\bar{a}_{T} \right) \right] \text{ and } \theta^b = \E\left[Y_{T+1}\left(\bar{b}_{T} \right) \right].
\end{equation}
\textbf{Objective:} We aim at estimating the average causal effect~(ACE) $\psi = \theta^a - \theta^b$. To do so, we impose three standard causal inference assumptions \cite{Robins.2000}: consistency, positivity, and sequential ignorability (see Appendix). Together, these assumptions allow to identify the ACE $\psi$ from observational data \cite{Pearl.2009b}. 

\subsection{Iterative G-computation}
\label{sec:iterative_gcomp}

In contrast to the static setting, simple covariate adjustment is not sufficient for identification of the ACE because the time-varying covariates may be caused by previous treatments \cite{Pearl.2009b}. The well-known \emph{G-formula} \cite{Robins.1986} offers a remedy by successively integrating out post-treatment covariates while conditioning on the intervention of interest. For our setting, we consider a variant that uses iterated conditional expectations \cite{Robins.1999,Bang.2005}. That is, we can write the parameter $\theta^a$ as the result of an iterative process. More precisely, we introduce recursively defined conditional expectations $Q_t^a$ that depend on the covariates $\bar{X}_{t-1}$ and interventions $\bar{a}_{t-1}$ via
\begin{equation}\label{eq:iterative_process}
   \resizebox{.85\hsize}{!}{$Q_t^{a} = Q_t\left(\bar{X}_{t-1}, \bar{a}_{t-1} \right) = \E\left[ Q_{t+1}^{a} \mid \bar{X}_{t-1}, \bar{A}_{t-1} = \bar{a}_{t-1} \right]$}
\end{equation}
for $t \in \{2, \dots, T+1\}$, and initialize $Q_{T+2}^{a} = Y_{T+1}$. Then, the expected potential outcome can be written as
\begin{equation}\label{eq:theta_iterativ}
    \theta^a = \E\left[Q_2^a \right].
\end{equation}
In Eq.~\eqref{eq:iterative_process}, the covariates $X_T, X_{T-1},\ldots$ are successively integrated out, and the $\theta^a$ is obtained in Eq.~\eqref{eq:theta_iterativ} by averaging. For a derivation, we refer to the Appendix.


In the following, we review iterative G-computation \cite{vanderLaan.2018}, which is an iterative procedure that leverages Eq.~\eqref{eq:iterative_process} and Eq.~\eqref{eq:theta_iterativ} to estimate the expected potential outcome $\theta^a$ and, subsequently, the average causal effect $\psi$. Iterative G-computation estimates the conditional expectations $Q_t^{a}$ by using a regression model for all $t \in \{2, \dots, T+1\}$. Then, $\theta^a$ can be estimated by taking the empirical mean in Eq.~\eqref{eq:theta_iterativ}. The full algorithm is in Alg.~\ref{alg:iter_gcomp}.

\vspace{-0.2cm}
\begin{algorithm}
\DontPrintSemicolon
\caption{Iterative G-computation \cite{Robins.1999, vanderLaan.2018}}
\scriptsize
\label{alg:iter_gcomp}
$\hat{Q}_{T+2}^a \gets Y_{T+1}$\;
\For{$t \in \{T+1,T,\dots,2\}$}{
$\hat{Q}_t(\cdot) \gets$ Regress $\hat{Q}_{t+1}^a$ on $\left(\bar{X}_{t-1}, \bar{A}_{t-1}\right)$\;
 $\hat{Q}_{t}^a \gets \hat{Q}_t\left(\bar{X}_{t-1}, \bar{a}_{t-1} \right)$\;
}
$\hat{\theta}^a \gets \frac{1}{N} \sum_{i=1}^N \hat{Q}_2^{a^{(i)}}$
\end{algorithm}
\vspace{-0.2cm}


Iterative G-computation in the above form is subject to drawbacks. In particular, one has to specify $T$ separate regression models that are trained separately. Each model uses the predictions of its already trained predecessor as training labels. This may lead to a training procedure which is comparatively unstable. 


\textbf{Need for an end-to-end model}: We postulate that an end-to-end model that learns all regression models jointly should overcome the above drawbacks. In particular, an end-to-end model can share information across time steps and, thereby, should be able to generate more accurate estimates for time-varying ACEs. Motivated by this, our end-to-end model learns all parameters jointly.

\section{DeepACE}


\textbf{Overview:} We propose a novel end-to-end deep learning model for ACE estimation, called DeepACE. DeepACE is motivated by the idea of iterative G-computation. It is trained with observational data from patient trajectories and a specific treatment intervention $\bar{a}_T$, based on which it learns the conditional expectations $Q_t^{a}$ from Eq.~\eqref{eq:iterative_process}. 

\begin{figure*}
\centering
\includegraphics[width=0.6\textwidth]{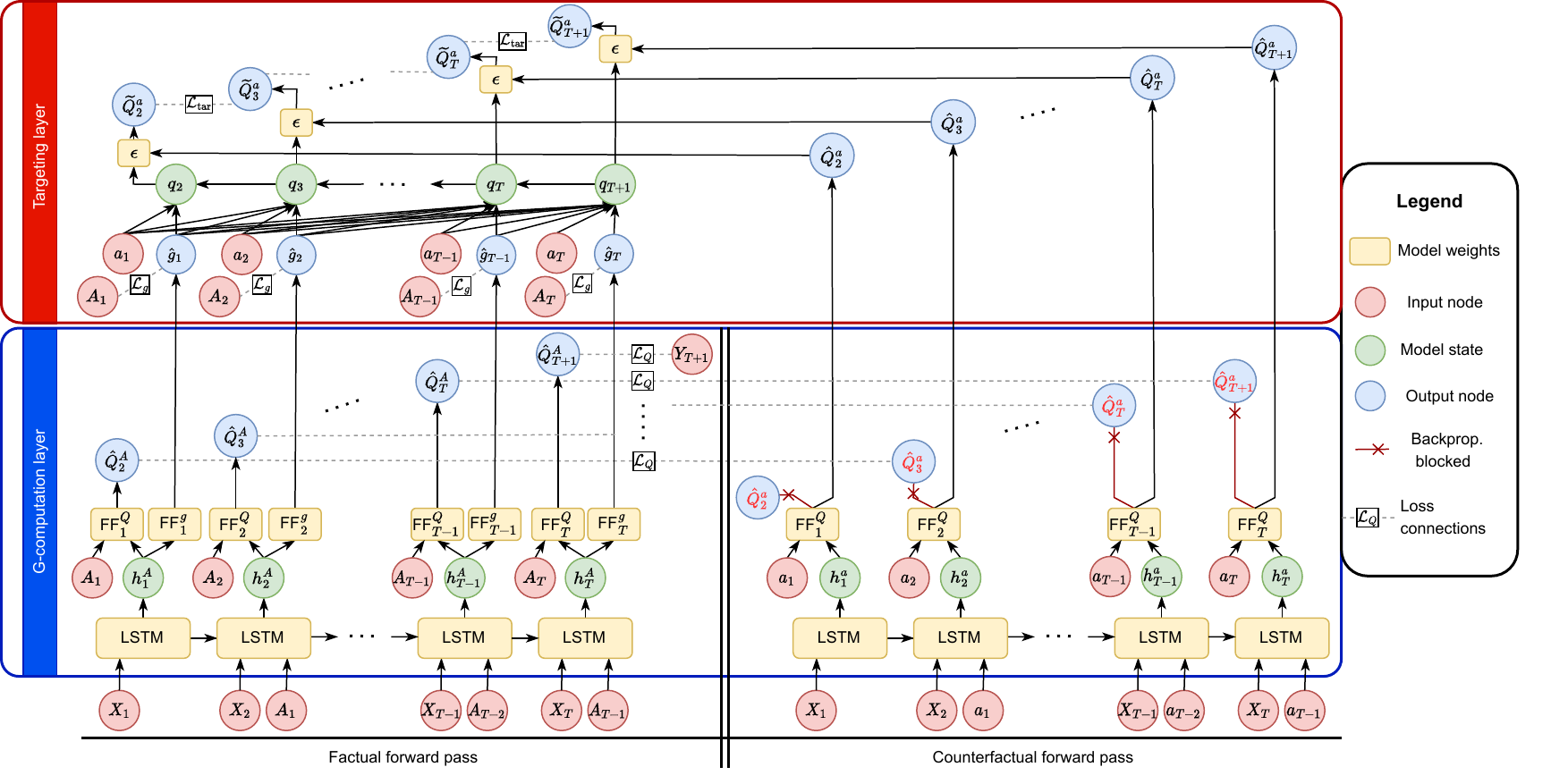}
\vspace{-0.2cm}
\caption{DeepACE consisting of the  \textcolor{blue}{\textbf{G-computation layer}} and the \textcolor{red}{\textbf{targeting layer}}.}
\label{fig:deepace}
\vspace{-0.5cm}
\end{figure*}

DeepACE consists of two main components: (i)~a \textbf{\textcolor{blue}{G-computation layer}} (Sec. G-computation layer), which produces initial estimates of the $Q_t^{a}$ by minimizing a tailored G-computation loss, and (ii)~a \textbf{\textcolor{red}{targeting layer}} (Sec. Targeting layer), which applies perturbations in a way that the final estimator satisfies an efficient estimating equation. The overall model architecture is shown in Fig.~\ref{fig:deepace}. 

DeepACE is trained by combining (i)~a \emph{G-computation loss} $\mathcal{L}_Q$, (ii)~a \emph{propensity loss} $\mathcal{L}_g$, and (iii)~a \emph{targeting loss} $\mathcal{L}_{\mathrm{tar}}$ into a joint loss $\mathcal{L}$ as described later. The outputs of DeepACE can then be used to provide ACE estimates (Sec. ACE estimation). We further show that DeepACE provides a doubly robust and asymptotically efficient estimator (Sec. Theoretical results). Finally, we provide implementation details are described in Sec. Model training. 

\subsection{G-computation layer}
\label{sec:g_layer}

The G-computation layer takes the observational data $\mathcal{D}$ and a specific treatment intervention $\bar{a}_T$ as input. For each time step $t \in \{2,\dots, T+1\}$, it generates two outputs: (\lowroman{1}) a factual output $\hat{Q}_t^A$ for $Q_t\left(\bar{X}_{t-1} \bar{A}_{t-1} \right)$, and (\lowroman{2}) a counterfactual output $\hat{Q}_t^a$ for $Q_t\left(\bar{X}_{t-1}, \bar{a}_{t-1} \right)$ according to Eq.~\eqref{eq:iterative_process}. The factual outputs $\hat{Q}_t^A$ are trained to estimate the one-step shifted counterfactual outputs $\hat{Q}_{t+1}^a$, while the counterfactual outputs $\hat{Q}_t^a$ are obtained by implicitly evaluating $\hat{Q}_t\left(\bar{X}_{t-1}, \cdot \right)$ at $\bar{A}_{t-1} = \bar{a}_{t-1}$ as done in Alg.~\ref{alg:iter_gcomp}.

\subsubsection{Architecture} 

The architecture of the \textbf{\textcolor{blue}{G-computation layer}} is shown in Fig.~\ref{fig:deepace} (bottom). In the G-computation layer, we use a long short-term-memory (LSTM) layer \cite{Hochreiter.1997} to process the input data. We choose an LSTM due to its ability to learn complex non-linear dynamics from patient trajectories while addressing the vanishing gradient problem which frequently occurs when using recurrent neural networks.

We feed the data twice into the LSTM: (i)~with the observed treatments $\bar{A}_T$ (factual forward pass), and (ii)~once with the treatment intervention $\bar{a}_T$ (counterfactual forward pass). Based on this, we computed the hidden LSTM states as follows. At each time step $t$, the factual forward pass leads to a factual hidden LSTM state $h_t^A$ depending on the factual trajectory $\mathcal{H}_t = (\bar{X}_t, \bar{A}_{t-1})$, and the counterfactual forward pass leads to a counterfactual hidden LSTM state $h_t^{a}$ depending on the past covariates $\bar{X}_t$ and interventions $\bar{a}_{t-1}$.

Both hidden states $h_t^A$ and $h_t^a$ are processed further. In the factual forward pass, we feed the factual hidden state $h_t^{A}$ together with the current observed treatment $A_t$ into a fully-connected feed-forward network $\mathrm{FF}_t^Q$. The network $\mathrm{FF}_t^Q$ generates a factual output $\hat{Q}_{t+1}^A$ for $Q_{t+1}(\bar{X}_{t}, \bar{A}_{t})$ according to Eq.~\eqref{eq:iterative_process}. In the counterfactual forward pass, we feed the counterfactual hidden state $h_t^{a}$ also $\mathrm{FF}_t^Q$ and replace the treatment input $A_t$ with the current intervention $a_t$. As a result, the network $\mathrm{FF}_t^Q$ generates a counterfactual output $\hat{Q}_{t+1}^a$ for $Q_t^a  = Q_{t+1}(\bar{X}_{t}, \bar{a}_{t}) $.

\subsubsection{G-computation loss} 

We design a tailored loss function, such that we mimic Algorithm~\ref{alg:iter_gcomp}. For this, we denote the outputs of the G-computation layer for a patient $i$ at time $t$ by $\hat{Q}_{t+1}^{A^{(i)}}(\eta)$ and $\hat{Q}_{t+1}^{a^{(i)}}(\eta)$. Here, we explicitly state the dependence on the model parameters (\ie, the LSTM and feed-forward layers), which we denote by $\eta$. We define the \emph{G-computation loss} as
\begin{equation}
      \resizebox{.8\hsize}{!}{$\mathcal{L}_Q(\eta) = \frac{1}{N}\frac{1}{T} \sum_{i=1}^N \sum_{t = 2}^{T+1} \left(\hat{Q}_{t}^{A^{(i)}}(\eta) - \hat{Q}_{t+1}^{a^{(i)}}(\eta)\right)^2$},
\end{equation}
where we defined $\hat{Q}_{T+2}^{a^{(i)}}(\eta) = y_{T+1}^{(i)}$.

\textbf{Blocking counterfactual backpropagation:} Each counterfactual output $\hat{Q}_{t+1}^{a}$ is used as a prediction objective by the previous factual output $\hat{Q}_{t}^{A}$. Recall that, in Algorithm~\ref{alg:iter_gcomp}, the counterfactual estimates $\hat{Q}_{t+1}^{a}$ are obtained only by evaluating the learned conditional expectation at $\bar{A}_t = \bar{a}_t$. Therefore, we only want the factual outputs $\hat{Q}_{t}^{A}$ to learn the counterfactual outputs $\hat{Q}_{t+1}^{a}$ and not vice-versa. Hence, when training the model with gradient descent, we block the gradient backpropagation through the counterfactual $\mathrm{FF}_t^Q$ during the counterfactual forward pass.

\subsection{Targeting layer}
\label{sec:t_layer}

For our sequential targeting procedure, we now introduce a \textcolor{red}{\textbf{targeting layer}}. The motivation is as follows: In principle, we could estimate the expected potential outcome $\theta^a$ by first training the G-computation layer, and subsequently following Eq.~\eqref{eq:theta_iterativ} and taking the empirical mean over the first counterfactual outputs $\hat{Q}_{2}^a$. Instead, we propose to leverage results from semi-parametric estimation theory, as this allows us to construct an estimator with better theoretical properties, namely double robustness and asymptotic efficiency.\footnote{For an overview on semi-parametric estimation theory, we refer to \cite{Kennedy.2016}.} For this purpose, we design our targeting layer so that it estimates the \emph{propensity scores} $g_t(\mathcal{H}_t) = P(A_t \,\mid\, \mathcal{H}_t)$.

We then use the propensity scores to perturb the counterfactual outputs $\hat{Q}_{t}^a$ to make them satisfy an efficient estimating equation. To formalize this, we first provide the mathematical background and subsequently describe how we implement the targeting layer.

\subsubsection{Mathematical background}


In the following, we summarize the general framework under which semi-parametric efficient estimators can be obtained. Let $\hat{Q}^{a} = (\hat{Q}_{2}^{a}, \dots, \hat{Q}_{T+1}^{a})$ be estimators of the conditional expectations $(Q_{2}^{a}, \dots, Q_{T+1}^{a})$ from Eq.~\ref{eq:iterative_process}, and let $\hat{g} = (\hat{g}_1, \dots, \hat{g}_{T})$  be estimators of the propensity scores $(g_1, \dots, g_T)$, where $g_t = g_t(\mathcal{H}_t)$. Furthermore, let $\hat{\theta}^a$ be an estimator of $\theta^a$.

Ideally, we would like to obtain a tuple of estimators $(\hat{Q}^{a}, \hat{g}, \hat{\theta}^a)$ with the following properties: (1)~\emph{Double robustness}: If either $\hat{Q}^{a}$ or $\hat{g}$ are consistent, $\hat{\theta}^a$ is a consistent estimator of $\theta^a$. (2)~\emph{(Semi-parametric) asymptotic efficiency}: If both $\hat{Q}^{a}$ and $\hat{g}$ are consistent, $\hat{\theta}^a$ achieves the smallest variance among all asymptotically linear estimators of $\theta^a$.
It can be shown that, asymptotically, the tuple $(\hat{Q}^{a}, \hat{g}, \hat{\theta}^a)$ fulfills properties (1) and (2) if it satisfies the following \emph{efficient estimating equation} \cite{Kennedy.2016}
\begin{equation}\label{eq:estimating_eq}
    \resizebox{.5\hsize}{!}{$\frac{1}{N} \sum_{i=1}^N \phi\left(\hat{Q}^{a^{(i)}}, \hat{g}^{(i)}, \hat{\theta}^a \right) = 0 $},
\end{equation}
where $\phi$ is the efficient influence function of $(\hat{Q}^{a}, \hat{g}, \hat{\theta}^a)$.
We call an estimator that satisfies Eq.~\eqref{eq:estimating_eq} ``{targeted}''. For the longitudinal setting, $\phi$ has a closed form (derived in \citet{vanderLaan.2012}) that is given by
\begin{equation}
\resizebox{.89\hsize}{!}{$\phi\left(Q^{a}, g, \theta^a \right) =  \left(Q_{2}^a - \theta^a\right)  + \sum_{t = 2}^{T+1} \left( Q_{t+1}^a - Q_{t}^a \right) \left( \prod_{\ell = 1}^{t-1} \frac{\mathbbm{1}(A_\ell = a_\ell)}{g_\ell(\mathcal{H}_\ell)} \right)$},    
\end{equation}
where we used the convention that $Q_{T+2}^a = Y_{T+1}$ and where $\mathbbm{1}(\cdot)$ denotes the indicator function.

\subsubsection{Implementation} 
We propose a sequential targeting layer to perturb the initial estimates produced by the G-computation layer in order to satisfy Eq.~\eqref{eq:estimating_eq}. Specifically, we add a model parameter that is jointly trained with the other model parameters. A tailored regularization term ensures that the efficient estimating estimation from Eq.~\eqref{eq:estimating_eq} is satisfied. 


Inputs to the targeting layer are (i)~the counterfactual outputs $\hat{Q}_{t+1}^{a}(\eta)$ of the G-computation layer and (ii)~predictions $\hat{g}_t(\eta)$ of the propensity scores $g_t(\mathcal{H}_t)$, where, $\eta$ denotes the trainable parameters of the G-computation layer. To allow for gradient backpropagation, we obtain the counterfactual outputs $\hat{Q}_{t+1}^{a}(\eta)$ from a second (identical) output of $\mathrm{FF}_t^Q$, where the gradient flow is not blocked during training. Furthermore, we generate the propensity estimates $\hat{g}_t(\eta)$ by adding separate feed-forward networks $\mathrm{FF}_t^g$ on top of the factual hidden states $h_t^A$.

In the following, we describe how the targeting layer applies perturbations to generate targeted outputs $\widetilde{Q}_{t+1}$. We recursively define perturbation values $q_{T+2}(\eta) = 0$ and
\begin{equation}
    \resizebox{.55\hsize}{!}{$q_{t}(\eta) = q_{t+1}(\eta) - \prod_{\ell = 1}^{t-1} \frac{\mathbbm{1}(A_\ell = a_\ell)}{\hat{g}_\ell(\eta)}$}
\end{equation}
for $t \in \{2,\dots, T+1\}$. The perturbation values are used to create the targeted network outputs via \begin{equation}
    \widetilde{Q}_{t}^{a}(\eta, \epsilon) = \hat{Q}_{t}^{a}(\eta) + \epsilon q_t(\eta)
\end{equation}
for $t \in \{2,\dots,T+2\}$, where $\epsilon$ is an additional network parameter that is trained together with $\eta$. Note that, by definition, we have that $\widetilde{Q}_{T+2}^{a}(\eta, \epsilon) = Y_{T+1}$.

\subsubsection{Loss} 

We use two regularization terms in order to train the targeting layer. First, we define the \emph{propensity loss}
\begin{equation}
      \resizebox{.7\hsize}{!}{$\mathcal{L}_g(\eta) = \frac{1}{N}\frac{1}{T} \sum_{i=1}^N \sum_{t = 1}^T \mathrm{BCE}\left(\hat{g}_{t}^{(i)}(\eta), a_{t}^{(i)}\right),$}
\end{equation}
where $\mathrm{BCE}$ denotes binary cross-entropy loss. The {propensity loss} ensures that the propensity networks learn to predict the propensity scores $g_t(\mathcal{H}_t)$. Second, we define our \emph{targeting loss}
\begin{equation}
      \resizebox{.8\hsize}{!}{$\mathcal{L}_{\mathrm{tar}}(\eta, \epsilon) = \frac{1}{N}\frac{1}{T} \sum_{i=1}^N  \sum_{t = 2}^{T+1} \left(\widetilde{Q}_{t+1}^{a^{(i)}}(\eta, \epsilon) - \widetilde{Q}_{t}^{a^{(i)}}(\eta, \epsilon)\right)^2.$}
\end{equation}
We show in Sec. Theoretical results that our targeting loss forces the outputs $\widetilde{Q}_{t+1}$ to satisfy the efficient estimating from Eq.~\eqref{eq:estimating_eq} and thus makes them ``targeted''.

In contrast to other sequential targeting methods (\eg, LTMLE \cite{vanderLaan.2012}) that apply targeting perturbations iteratively over time, our procedure allows the entire model to be learned jointly. We later show that this gives more accurate ACE estimates. 

\subsection{Model training and ACE estimation}
\label{sec:ace_estimation}

\textbf{Overall loss:} To train DeepACE, we combine the above into an overall loss
\begin{equation}
    \mathcal{L}(\eta, \epsilon) = \mathcal{L}_Q(\eta) + \alpha \mathcal{L}_g(\eta) + \beta \mathcal{L}_{\mathrm{tar}}(\eta, \epsilon),     
\end{equation}
where $\alpha$ and $\beta$ are constants that control the amount of propensity and targeting regularization, respectively. Details on our implementation, training, and hyperparameter tuning are in the Appendix. 

\textbf{ACE estimation:} Given two treatment interventions $\bar{a}_T$ and $\bar{b}_T$, we train two separate DeepACE models for each $\bar{a}_T$ and $\bar{b}_T$. Then, we estimate the ACE $\psi$ via $\hat{\psi} = \frac{1}{N} \sum_{i=1}^N \left(\widetilde{Q}_{2}^{a^{(i)}} - \widetilde{Q}_{2}^{b^{(i)}}\right)$,
where $\widetilde{Q}_{2}^{a^{(i)}}$ and $\widetilde{Q}_{2}^{b^{(i)}}$ denote the two targeted DeepACE outputs for patient $i$ at time $t=2$.

\subsection{Theoretical results}
\label{sec:theory}

The following theorem ensures that our combination of targeting layer and regularization in DeepACE indeed produces a targeted estimator.
\newtheorem{thrm}{Theorem}
\begin{thrm}\label{thrm:target}
Let $(\hat{\eta}, \hat{\epsilon})$ be a stationary point of $\mathcal{L}(\eta, \epsilon)$. Then, for any $\beta > 0$, the estimator $\widetilde{\theta}^a = \frac{1}{N} \sum_{i=1}^N \widetilde{Q}_{2}^{a^{(i)}}(\hat{\eta}, \hat{\epsilon})$
is targeted, \ie, DeepACE satisfies the efficient estimating equation from Eq.~\eqref{eq:estimating_eq}.
\end{thrm}
\begin{proof}
See Appendix.
\end{proof}

\noindent
By Theorem~\ref{thrm:target}, the DeepACE estimator $\widetilde{\theta}^a$ is doubly robust, \ie, $\widetilde{\theta}^a$ is consistent, even if either the targeted outputs $\widetilde{Q}_{t+1}^a$ or the propensity estimates $\hat{g}_t$ are misspecified. 

Assuming that the true conditional expectations $Q_{t+1}^a$ and propensity scores $g_t$ are contained in a suitable hypothesis class, the initial outputs $\hat{Q}_{t+1}^a$ from the G-computation layer and the propensity estimates $\hat{g}$ will converge to $Q_{t+1}^a$ and $g_t$ with growing sample size due to the construction of $\mathcal{L}_Q$ and $\mathcal{L}_g$ (provided that $\alpha > 0$). The next corollary shows that this implies asymptotic efficiency of $\widetilde{\theta}^a$.

\newtheorem{cor}{Corollary}
\begin{cor}\label{cor:efficiency}
If the initial outputs $\hat{Q}_{t+1}^a$ from the G-computation layer and propensity estimates $\hat{g}_t$ of DeepACE are consistent for $Q_{t+1}^a$ and $g_t$, then also the targeted outputs $\widetilde{Q}_{t+1}^a$ are consistent for $Q_{t+1}^a$. In particular, the DeepACE estimator $\widetilde{\theta}^a$ is asymptotically efficient for $\theta^a$.
\end{cor}
\begin{proof}
See Appendix.
\end{proof}

\section{Experiments}
\label{sec:experiments}

\subsection{Baselines}

We compare DeepACE against state-of-the-art methods for time-varying causal effect estimation, see Table~\ref{t:rel_work}. The baselines are selected from recent literature on causal effect estimation \cite{Li.2021, vanderLaan.2012}. The baselines can be categorized into three groups: (1)~\textbf{G-methods}, (2)~Longitudinal targeted maximum likelihood estimation (\textbf{LTMLE}), and (3)~\textbf{deep learning models for ITEs}. Implementation and hyperparameter tuning details for all baselines are in the Appendix. 

\subsection{Experiments using synthetic data}\label{sec:exp_syn}

\textbf{Setting:}
Synthetic data are commonly used to evaluate the effectiveness of causal inference methods because they provide access to the counterfactual outcomes (\eg, \cite{vanderLaan.2006, Bica.2020b, Shi.2019}). Therefore, we can successfully compute the ground-truth ACE and thus benchmark the performance of all methods.

\textbf{Results:}
We generate a synthetic dataset with $N=1000$ patient trajectories over $T=15$ time steps and sample $p=6$ time-varying covariates. We then evaluate our baselines on three different setups which correspond to different treatment interventions \footnote{Here, $\bar{b}_T$ is fixed to the zero-intervention (no treatment applied), and $\bar{a}_T$ is chosen as $(\Indicator({k \leq i\leq \ell}))_{i \in \{1,\dots,T\}}$ for $k \in \{1,3,5\}$ and $\ell \in \{10, 13, 15\}$.}. For each method and setup, we calculate the absolute error between estimated and ground-truth averaged over 5 different runs with random seeds. We refer to the Appendix for details regarding the data generating process and method evaluation.

The results are shown in Table~\ref{t:results}. All baselines are clearly outperformed by DeepACE on all three experiments. The best-performing baseline is LTMLE with the super learner. This is reasonable as it is the only baseline available that is both tailored for ACE estimation and makes use of machine learning. The linear methods (\ie, g-methods, LTMLE, and glm) are not able to capture the non-linear dependencies within the data and thus achieve an inferior performance. The ITE baselines (except CT) use recurrent neural networks and should thus be able to learn non-linearities but, nevertheless, are inferior. This is unsurprising and attributed to the fact that they are designed for estimating individual rather than average causal effects (see Section Related work).

\begin{table}[ht]
\vspace{-0.2cm}
\caption{Results on synthetic data (mean $\pm$ std. dev.).}
\vspace{-0.3cm}
\centering
\label{t:results}

\resizebox{\columnwidth}{!}{\begin{tabular}{lccc}

\noalign{\smallskip} \toprule \noalign{\smallskip}
Method & Setup 1 & Setup 2 & Setup 3 \\
\midrule
\textsc{(1) g-methods} & &  & \\
\quad MSM \cite{Robins.2000} & $0.24 \pm 0.18$& $ 0.27\pm 0.19 $ & $0.21 \pm 0.10$\\
\quad Iterative G-computation \cite{vanderLaan.2018}&$0.26 \pm 0.23$ & $ 0.34\pm 0.26$& $0.32 \pm 0.27$\\
\quad G-formula (parametric) \cite{Robins.2009}&$0.14 \pm 0.10$ & $ 0.44\pm 0.25 $& $0.86 \pm 0.45$ \\
\quad SNMM \cite{Robins.1994}&$0.39 \pm 0.04$ & $ 0.47\pm 0.02 $& $0.35 \pm 0.02$\\
\midrule
\textsc{(2) LTMLE} & &  & \\
\quad LTMLE (glm) \cite{vanderLaan.2012} &$0.20 \pm 0.19$& $0.25 \pm 0.19$& $0.33 \pm 0.24$ \\
\quad LTMLE (super learner) \cite{vanderLaan.2012} & $0.13\pm 0.08$ & $0.14 \pm 0.12$  &$ 0.19\pm 0.13$\\
\midrule
\textsc{(3) Deep learning for ITE estimation} & &  & \\
\quad RMSNs \cite{Lim.2018}& $0.20 \pm 0.10$& $0.52 \pm 0.32$ & $0.95 \pm 0.52$\\
\quad CRN \cite{Bica.2020}& $0.20 \pm 0.10$ & $0.52 \pm 0.32$ & $0.95 \pm 0.52$\\
\quad G-Net \cite{Li.2021}&$0.18 \pm 0.08$& $0.46 \pm 0.26$ & $0.97 \pm 0.38$ \\
\noalign{\smallskip} \bottomrule \noalign{\smallskip}
DeepACE w/o targeting (ours) & $\boldsymbol{0.04 \pm 0.03}$ & $0.10 \pm 0.09$ & $\boldsymbol{0.15 \pm 0.11}$\\
DeepACE (ours) & $\boldsymbol{0.04 \pm 0.02}$ & $\boldsymbol{0.09 \pm 0.07}$ & $0.18 \pm 0.08$\\
\bottomrule
\multicolumn{4}{l}{lower $=$ better (best in bold)}
\end{tabular}}
\vspace{-0.4cm}
\end{table}

To compare DeepACE with iterative G-computation, we compute the estimation errors of both methods over different time lags $h \in \{1,\dots,5\}$, corresponding to level of long-range dependencies within the data (see Appendix). The results are shown in Fig.~\ref{fig:pltlag}, showing the effectiveness of DeepACE for long-range dependencies (common in EHRs).
\vspace{-0.3cm}
\begin{figure}[h]
\centering
\includegraphics[width=0.6\linewidth]{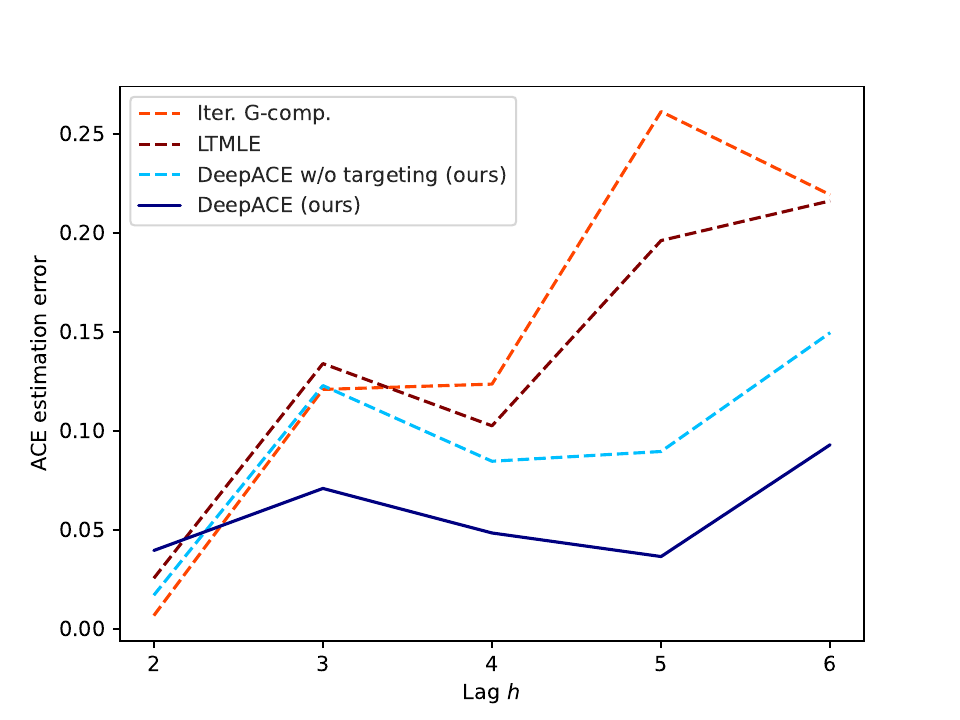}
\vspace{-0.2cm}
\caption{Performance comparison. Shown: mean estimation error averaged over 5 runs.}
\label{fig:pltlag}
\vspace{-0.4cm}
\end{figure}

\textbf{Ablation study:} We also analyze DeepACE but where the targeting layer is removed (Table~\ref{t:results}). Both variants of DeepACE outperform the baselines. The variant without targeting performs well but we are careful with interpretations due to the simple nature of the synthetic data and rather relegate conclusions to real-world data as in the following. 

\subsection{Experiments using semi-synthetic data}

\textbf{Setting:} We create semi-synthetic data that enables us to evaluate DeepACE using real-world data while having access to the ground-truth ACE. For this purpose, we use the MIMIC-III dataset \cite{Johnson.2016}, which includes electronic health records from patients admitted to intensive care units. We generate $N=1000$ patient trajectories. Again, compare three setups with different treatment interventions. For details, we refer to the Appendix.

\textbf{Results:} The results are shown in Table~\ref{t:results_mimic}. Again, DeepACE outperforms all baselines by a large margin.

\begin{table}[ht]
\vspace{-0.3cm}
\caption{Results on semi-synthetic data (mean $\pm$ std. dev.).}
\vspace{-0.3cm}
\centering
\label{t:results_mimic}
\scriptsize
\resizebox{\columnwidth}{!}{\begin{tabular}{lccc}
\noalign{\smallskip} \toprule \noalign{\smallskip}
{Method} & {Setup 1} & {Setup 2} & {Setup 3} \\
\midrule
\textsc{(1) g-methods} & &  & \\
\quad MSM \cite{Robins.2000} & $2.35 \pm 0.64$&$3.06 \pm 0.67$  &$2.47 \pm 0.95$ \\
\quad Iterative G-computation \cite{vanderLaan.2018}&$0.81 \pm 0.35$ & $0.72 \pm 0.72$& $1.90 \pm 1.06$\\
\quad G-formula (parametric) \cite{Robins.2009}&$0.32 \pm 0.27$ &$0.31 \pm 0.20$ & $0.32 \pm 0.27$  \\
\quad SNMM \cite{Robins.1994}& $0.28 \pm 0.33$&$0.52 \pm 0.26$ & $1.96 \pm 2.36$ \\
\midrule
\textsc{(2) LTMLE} & &  & \\
\quad LTMLE (glm) \cite{vanderLaan.2012}& $0.82 \pm 0.35$& $0.72 \pm 0.72$ & $1.84 \pm 1.13$ \\
\quad LTMLE (super learner) \cite{vanderLaan.2012} & $0.96 \pm 1.01$&  $0.92 \pm 1.17 $&$0.76 \pm 0.47$\\
\midrule
\textsc{(3) Deep learning for ITE estimation} & &  & \\
\quad RMSNs \cite{Lim.2018}& $2.35 \pm 0.14$ & $2.32 \pm 0.18$ & $2.36 \pm 0.14$\\
\quad CRN \cite{Bica.2020}& $2.53 \pm 0.03$& $2.53 \pm 0.04$ &$2.52 \pm 0.04$ \\
\quad G-Net \cite{Li.2021}& $0.67 \pm 0.15$& $0.65 \pm 0.17$ & $0.67 \pm 0.15$ \\
\noalign{\smallskip} \bottomrule \noalign{\smallskip}
DeepACE w/o targeting (ours) & $\boldsymbol{0.18 \pm 0.17}$ & $0.25 \pm 0.11$ & $0.21 \pm 0.07$\\
DeepACE (ours) & $\boldsymbol{0.18 \pm 0.14}$& $\boldsymbol{0.12 \pm 0.14}$ & $\boldsymbol{0.16 \pm 0.10}$\\
\bottomrule
\multicolumn{4}{l}{lower $=$ better (best in bold)}
\end{tabular}}
\vspace{-0.4cm}
\end{table}

\textbf{Ablation study:} We repeat the experiments with DeepACE but where the targeting layer is removed (Table~\ref{t:results_mimic}). This thus demonstrates the importance of our targeting procedure for achieving a superior performance.

\subsection{Case study using real-world data}

\textbf{Setting:} We demonstrate the value of DeepACE for real-world patient trajectories collected in a clinical study. For this purpose, we analyze data from $N = 928$ patients with low back pain (LBP) \cite{Nielsen.2017}. Here, we are interested in the causal effect of whether patients have been allowed/disallowed to do physical labor (binary treatments) on pain intensity (outcome). Thus, the treatment interventions of interest are $a = (1, 1, 1)$ (physical labor) and $b = (0, 0, 0)$ (stop physical labor) with $T=3$. Medical research is interested in identifying different phenotypes whereby LBP is classified into different subgroups with clinical meaningful interpretation. We thus estimate average causal effects $\psi_i = \theta^a_i - \theta^b_i$ for two patient cohorts, namely (1)~severe LBP ($i=1$) and (2)~mild LBP ($i=2$) \cite{Nielsen.2017}.

\vspace{-0.4cm}
\begin{figure}[ht]
\centering
\includegraphics[width=0.55\linewidth]{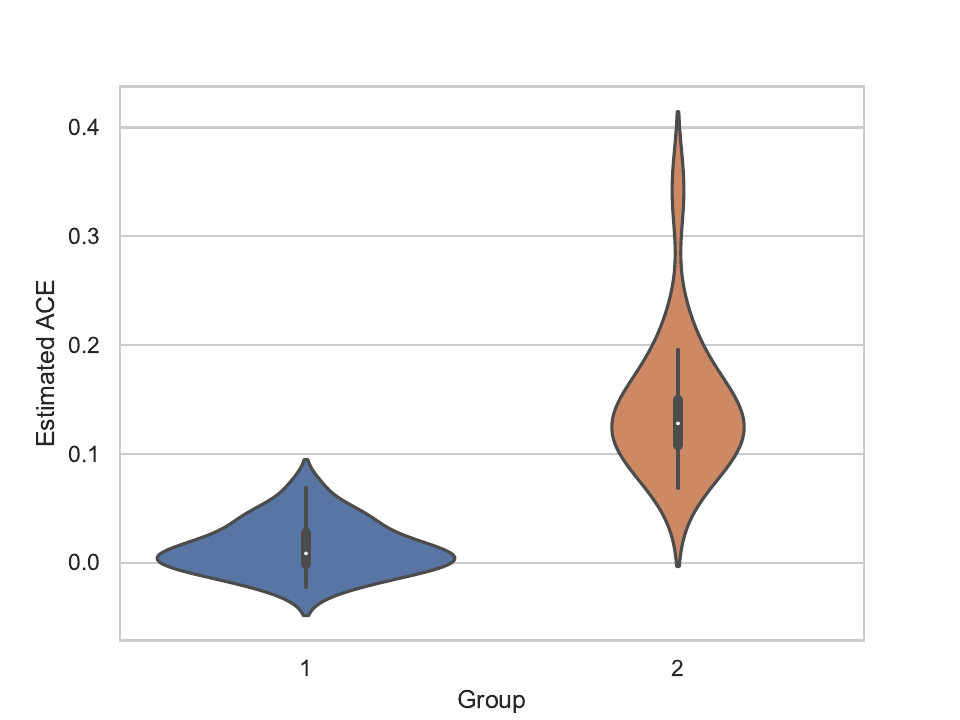}
\vspace{-0.2cm}
\caption{Violin charts (boxplots) showing the distributions of DeepACE estimates $\psi_i$ (standardized) for two patient subgroups $i \in \{1, 2\}$ The distributions are obtained using Monte Carlo dropout during model prediction.}
\label{fig:plt_backpain}
\vspace{-0.2cm}
\end{figure}

\textbf{Results:} We make several observations relevant for medical practice (Fig.~\ref{fig:plt_backpain}): (i)~Both ACEs are positive, which is line with medical knowledge implying that physical labor may worsen LBP progression. (ii)~Surprisingly, the ACE is larger for the group with mild LBP. Hence, our results suggest that physical labor should be also stopped for mild LBP, as this would eliminate negative effects for observed pain. (iii)~The variance is much larger for mild LBP, indicating that impact of physical labor is more heterogeneous.  

\section{Discussion}

Estimating causal effects from observational data requires custom methods that adjust for time-varying confounding. Recently proposed deep learning models aim at estimating time-varying \emph{individual treatment effects} (ITEs) and are not able to provide efficient \emph{average causal effect} (ACE) estimators. DeepACE fills this gap by improving on state-of-the-art methods for time-varying ACE estimation.

\textbf{Future work:} Our work could be extended in several ways. Two possible directions are: (1)~DeepACE could be extended to account for dynamic treatment regimes \cite{Robins.2009} instead of prespecified static ones. (2)~The performance of targeted estimation under violation of identification assumptions (ignorability, positivity) could be of interest.

\bibliography{bibliography}

\clearpage

\appendix

\section{Proofs}\label{sec:appendix_proofs}

\begin{proof}[Proof of Theorem \ref{thrm:target}]
We show that the tuple $\left(\widetilde{Q}^{a}(\hat{\eta}, \hat{\epsilon}), \bar{g}(\hat{\eta}), \widetilde{\theta}^a\right)$ satisfies the efficient estimating equation in Eq.~\eqref{eq:estimating_eq}. At any stationary point $(\hat{\eta}, \hat{\epsilon})$, we have that
{\footnotesize%
\begin{align}
    0 &= \frac{\partial}{\partial \epsilon} \mathcal{L}(\eta, \epsilon) \Bigr|_{\substack{(\eta, \epsilon) = (\hat{\eta}, \hat{\epsilon})}} \\
    &= \beta \frac{\partial}{\partial \epsilon} \mathcal{L}_{\mathrm{tar}}(\eta, \epsilon) \Bigr|_{\substack{\substack{(\eta, \epsilon) = (\hat{\eta}, \hat{\epsilon})}}} \\
    &= \frac{\beta}{NT} \sum_{i=1}^N  \sum_{t = 2}^{T+1} \left(\widetilde{Q}_{t+1}^{a^{(i)}}(\hat{\eta}, \hat{\epsilon}) - \widetilde{Q}_{t}^{a^{(i)}}(\hat{\eta}, \hat{\epsilon})\right) \left(q_{t+1}(\hat{\eta}) - q_t(\hat{\eta}) \right) \\
    &= \frac{\beta}{NT} \sum_{i=1}^N  \sum_{t = 2}^{T+1} \left(\widetilde{Q}_{t+1}^{a^{(i)}}(\hat{\eta}, \hat{\epsilon}) - \widetilde{Q}_{t}^{a^{(i)}}(\hat{\eta}, \hat{\epsilon})\right) \prod_{\ell = 1}^{t-1} \frac{\mathbbm{1}(a_\ell^{(i)} = a_\ell)}{\hat{g}_\ell(\hat{\eta})} \\
    &= \frac{\beta}{NT} \sum_{i=1}^N \phi\left(\bar{Q}_{T+1}^{a}(\hat{\eta}, \hat{\epsilon}), \widetilde{g}(\hat{\eta}), \widetilde{\theta}^a \right).
\end{align}}
Multiplying both sides with $\frac{T}{\beta}$ yields the result.
\end{proof}

\begin{proof}[Proof of Corollary \ref{cor:efficiency}]
We show that consistency of $\hat{Q}_{t+1}^a$ and $\hat{g}_t$ implies consistency of the targeted outputs $\widetilde{Q}_{t+1}^a$. Then, asymptotic efficiency follows from Theorem~\ref{thrm:target}.

The key argument is similar as in \cite{Shi.2019}. By assumption, the G-computation loss $\mathcal{L}_Q$ and the propensity loss $\mathcal{L}_g$ are asymptotically minimized by $\hat{Q}_{t+1}^a = Q_{t+1}^a$ and $\hat{g}_t = g_t$ (due to finite Vapnik–Chervonenkis (VC) dimension). In the following, we show that the overall loss $\mathcal{L}$ is asymptotically minimized by additionally setting $\hat{\epsilon} = 0$. Hence, the targeted outputs $\widetilde{Q}_{t+1}^a$ are consistent because DeepACE can set the perturbation parameter to $\hat{\epsilon} = 0$, which implies $\widetilde{Q}_{t+1}^a = \hat{Q}_{t+1}^a$ for all $t \in \{1,\dots, T\}$.

Because the targeting layer only adds a single parameter to the model, a finite VC dimension is preserved. We show now that $\hat{\epsilon} = 0$ asymptotically minimizes each summand of the targeting loss $\mathcal{L}_{\mathrm{tar}}$. The last summand for $t = T+1$ is minimized (squared loss) at
\begin{equation}
    \hat{Q}_{T+1}^a + \hat{\epsilon} q_{T+1} = \E [Y_{T+1} \mid \bar{X}_T, \bar{A}_T = \bar{a}_T] = Q_{T+1}^a,
\end{equation}
which is indeed achieved at $\hat{\epsilon} = 0$. The other summands for $t \in \{2, \dots, T\}$ are minimized at
\begin{align}
        \hat{Q}_{t}^a + \hat{\epsilon} q_{t} &= \E[\widetilde{Q}_{t+1}^a \mid \bar{X}_{t-1}, \bar{A}_{t-1} = \bar{a}_{t-1}] \\
        &= Q_{t}^a + \hat{\epsilon} \E[q_{t+1}^a \mid \bar{X}_{t-1}, \bar{A}_{t-1} = \bar{a}_{t-1}],
\end{align}
which is also achieved at $\hat{\epsilon} = 0$.
\end{proof}

\section{Causal inference assumptions and identification}
\label{sec:app:assumptions}

Throughout the paper, we impose the following assumptions on the data-generating process. These assumptions are standard in causal inference literature \cite{Robins.2000, Lim.2018, Bica.2020b}.

\newtheorem{assumption}{Assumption}
\begin{assumption}[Consistency]\label{ass:consistency}
The treatment assignment $\bar{A}_{t} = \bar{a}_{t}$ implies $Y_{t+1}\left(\bar{a}_{t} \right) = Y_{t+1}$, \ie, potential and observed outcomes coincide for all $t \in \{1, \ldots, T\}$.
\end{assumption}
\begin{assumption}[Positivity]\label{ass:positivity}
If $P(\bar{A}_{t-1} = \bar{a}_{t-1}$, $\bar{X}_t= \bar{x}_t) > 0$, then $P(A_t = a_t \,\mid\, \bar{A}_{t-1} = \bar{a}_{t-1}, \bar{X}_t= \bar{x}_t) > 0$ for all $a_t \in \{0,1\}$.
\end{assumption}
\begin{assumption}[Sequential ignorability]\label{ass:seq_ignor} 
For all $\bar{a}_t \in \{0,1\}^t$ and $t \in \{1, \ldots, T\}$, it holds that
$Y_{t+1}(\bar{a}_t)) \indep A_t \,\mid\, \mathcal{H}_t$.
\end{assumption}

\noindent
Under Assumption~\ref{ass:positivity}, the potential outcome associated with a treatment intervention that has been observed coincides with the observed outcome. In particular, this implies that stable unit treatment value assumption (SUTVA): An intervention on a specific patient does not influence the outcome of another patient. Assumption~\ref{ass:positivity} states that almost surely, there is a nonzero probability of giving treatment and not giving treatment conditional on an arbitrary patient history. 
Assumption~\ref{ass:seq_ignor} implies that there are no unobserved confounders that influence both treatment assignments and outcomes. 

Together, Assumptions~\ref{ass:consistency}--\ref{ass:seq_ignor} allow us to identify the expected potential outcomes (and thus the ACE $\psi$) from observational data via the well-known G-formula \cite{Pearl.2009b, Robins.1986}. We make this precise in the following Theorem. 

\begin{thrm}\label{thrm:gcomp}
Under Assumptions 1-3, the expected potential outcome $\theta^a$ is identified as
\begin{equation}\label{eq:g_comp}
\begin{split}
    \theta^a = & \E\left[ \E\left[ \ldots \; \E\left[ \E\left[ Y_{T+1} \;\middle|\; \bar{X}_T, \bar{A}_T = \bar{a}_T  \right] \; \right. \right. \right. \\ 
    & \left. \left. \left. \middle| \bar{X}_{T-1}, \bar{A}_{T-1} = \bar{a}_{T-1} \right] \ldots \; \middle|\; X_1, A_1 = a_1 \right] \right]
\end{split}
\end{equation}
for any treatment intervention $\bar{a}_{T} = (a_1,\dots,a_{T})$.
\end{thrm}
\begin{proof}
It holds that
\begin{equation}
\begin{split}
    \theta^a &=  \E\left[ Y_{T+1}(\bar{a}_{T}) \right] \\
    &\overset{(1)}{=} \E\left[\E\left[ Y_{T+1}(\bar{a}_{T}) \;\middle|\; X_1 \right] \right] \\
    &\overset{(2)}{=} \E\left[\E\left[ Y_{T+1}(\bar{a}_{T}) \;\middle|\; X_1, A_1 = a_1 \right] \right] \\
    &\overset{(3)}{=} \E\left[\E\left[ \E\left[Y_{T+1}(\bar{a}_{T}) \;\middle|\; \bar{X}_2, A_1 = a_1 \right] \;\middle|\; X_1, A_1 = a_1 \right] \right] \\
    &\overset{(4)}{=} \E\left[\E\left[ \E\left[Y_{T+1}(\bar{a}_{T}) \;\middle|\; \bar{X}_2, \bar{A}_2 = \bar{a}_2 \right] \;\middle|\; X_1, A_1 = a_1 \right] \right] \\
    &= \ldots \\
    &= \E\left[ \E\left[ \ldots \; \E\left[ \E\left[ Y_{T+1}(\bar{a}_{T}) \;\middle|\; \bar{X}_T, \bar{A}_T = \bar{a}_T  \right] \right. \right. \right. \\
    & \left. \left. \left. \quad  \middle| \bar{X}_{T-1}, \bar{A}_{T-1} = \bar{a}_{T-1} \right] \ldots \; \middle|\; X_1, A_1 = a_1 \right] \right] \\
    &\overset{(5)}{=}\E\left[ \E\left[ \ldots \; \E\left[ \E\left[ Y_{T+1} \;\middle|\; \bar{X}_T, \bar{A}_T = \bar{a}_T  \right] \right. \right. \right. \\
    & \left. \left. \left. \quad \middle| \bar{X}_{T-1}, \bar{A}_{T-1} = \bar{a}_{T-1} \right] \ldots \; \middle|\; X_1, A_1 = a_1 \right]  \right].
\end{split}
\end{equation}
Here, (1) follows from the law of total expectation, (2) from Assumption~\ref{ass:positivity} and ~\ref{ass:seq_ignor}, (3) from the tower property, (4) from Assumption~\ref{ass:positivity} and ~\ref{ass:seq_ignor}, and (5) from Assumption~\ref{ass:consistency}.
\end{proof}

As a consequence of Theorem~\ref{thrm:gcomp} we obtain the recursion formula Eq.~\eqref{eq:gcomp_recursion} for $\theta^a$ which is the basis of iterative G-computation.

\begin{cor}
We define the conditional expectations $Q_t^a$ that depend on the covariates $\bar{X}_{t-1}$ and interventions $\bar{a}_{t-1}$ via
\begin{equation}\label{eq:gcomp_recursion}
  Q_t^{a} = Q_t\left(\bar{X}_{t-1}, \bar{a}_{t-1} \right) = \E\left[ Q_{t+1}^{a} \mid \bar{X}_{t-1}, \bar{A}_{t-1} = \bar{a}_{t-1} \right]
\end{equation}
for $t \in \{2, \dots, T+1\}$, and initialize $Q_{T+2}^{a} = Y_{T+1}$. Then, the expected potential outcome can be expressed as
\begin{equation}
    \theta^a = \E\left[Q_2^a \right].
\end{equation}
\end{cor}
\begin{proof}
This follows directly from Theorem~\ref{thrm:gcomp}.
\end{proof}

\section{Implementation of baselines}\label{sec:appendix_baseline}

We provide a detailed overview of our baselines. Of note, our baselines represent state-of-the-art methods for causal effect estimation in longitudinal settings \cite{Bica.2020, Lim.2018, vanderLaan.2012, Li.2021}.

(1) \textbf{G-methods:} Here, we use: (i)~a marginal structural network (\textbf{MSN}) \cite{Robins.2000} with inverse probability weighting, (ii)~\textbf{iterative G-computation} as in Algorithm~\ref{alg:iter_gcomp}, (iii) \textbf{G-computation} via the parametric G-formula \cite{Robins.2009}, and (iv)~a structural nested mean model (\textbf{SNMM}) with g-estimation \cite{Robins.1994}.

(2) \textbf{Longitudinal targeted maximum likelihood estimation} (\textbf{LTMLE}) \cite{vanderLaan.2012}: We implement LTMLE in two variants. The first variant~(i) uses generalized linear models \textbf{(glm)} to estimate the conditional expectations. The second variant~(ii) uses the \textbf{super learner} algorithm \cite{vanderLaan.2007}, which builds upon a cross-validation algorithm to combine different regression models into a single predictor.

(3) \textbf{Deep learning for ITE estimation:}
Additionally, we include (i)~recurrent marginal structural networks (\textbf{RMSNs}) \cite{Lim.2018}, (ii)~a counterfactual recurrent network (\textbf{CRN}) \cite{Bica.2020}, and (iii)~\textbf{G-Net} \cite{Li.2021}. Different from our method, these baselines predict individual as opposed to average causal effects. We obtain ACE estimates by averaging the predicted ITEs.

In the following, we provide implementation details on all baselines. Hyperparameter tuning details are in Appendix~\ref{sec:appendix_hyper}. 

\subsection{G-methods}

Generalized methods (g-methods) \cite{Naimi.2017} are a class of statistical models for time-varying ACE estimation originally used in epidemiology \cite{Robins.2000}. G-methods can be loosely categorized into marginal structural models, G-computation via the G-formula, and structural nested models. We use models of all three categories.

\subsubsection{Marginal structural model (MSM) \cite{Robins.2000}}

MSMs express the expected potential outcome $\theta^a$ directly as a function of the treatment intervention $\bar{a}_T = (a_1, \dots, a_T)$. In our case, we consider the model
\begin{equation}
    \theta^a = \beta_0 + \sum_{t=1}^T \beta_t a_t
\end{equation}
with parameters $\beta = (\beta_0, \dots, \beta_T)$. The parameters $\beta$ can be estimated via inverse probability weighting. More precisely, the stabilized inverse probability weight for patient $i$ is defined as
{\footnotesize%
\begin{equation}
\label{eq:sw}
    SW_i = \frac{\prod_{t=1}^T f_{A_t \vert \bar{A}_{t-1}}\left(A_t^{(i)}, \bar{A}_{t-1}^{(i)}\right)}{\prod_{t=1}^T g_{A_t \vert \mathcal{H}_t}\left(A_t^{(i)}, \bar{A}_{t-1}^{(i)},\bar{X}_{t}^{(i)} \right)},
\end{equation}}
where $f_{A_t \vert \bar{A}_{t-1}}$ and $g_{A_t \vert \mathcal{H}_t}$ denote the conditional densities of $A_t$ given $\bar{A}_{t-1}$ and $\mathcal{H}_t = (\bar{A}_{t-1}, \bar{X}_{t})$, respectively. Both conditional densities can be estimated via standard logistic regressions.

Robins \cite{Robins.2000} showed that the parameters $\beta$ of the MSN can be consistently estimated by performing a weighted linear regression. Here, the observed treatments $A$ are regressed on the observed outcomes $Y_{T+1}$ and each observation $i$ is weighted by $SW_i$. Once the estimates $\hat{\beta}$ are obtained, we estimate the ACE via
\begin{equation}
    \hat{\psi} = \sum_{t=1}^T \hat{\beta}_t (a_t - b_t),
\end{equation}
where $\bar{a}_T$ and $\bar{b}_T$ denote the treatment interventions.

\subsubsection{G-computation \cite{Robins.2009, vanderLaan.2018}}

The G-formula from Eq.~\eqref{eq:g_comp} can be used to estimate $\theta^a$ in an iterative manner. We include Algorithm~\ref{alg:iter_gcomp} as a baseline in which we use linear regression to estimate the conditional expectations.

An equivalent way to write the G-formula from Eq.~\eqref{eq:g_comp} is
{\footnotesize%
\begin{equation}\label{eq:g_comp_par}
\begin{split}
        \theta^a = \int_{\R^{p \times \dots \times p}} &  \E\left[Y_{T+1} \;\middle|\; \bar{X}_T = \bar{x}_T, \bar{A}_T = a \right]  \\ & \prod_{t=1}^T  f_{X_t \vert \bar{X}_{t-1}, \bar{A}_{t-1}} \left(x_t \mid \bar{x}_{t-1}, \bar{a}_{t-1} \right) \; \mathrm{d}\bar{x}_T,
\end{split}
\end{equation}}
where $f_{X_t \vert \bar{X}_{t-1}, \bar{A}_{t-1}}$ denotes the conditional density of $X_t$ given $\bar{X}_{t-1}$ and $\bar{A}_{t-1}$. The conditional densities and the conditional expectation $\E\left[Y_{T+1} \;\middle|\; \bar{X}_T = \bar{x}_T, \bar{A}_T = a \right]$ can be estimated by parametric regression models and are subsequently plugged into Eq.~\eqref{eq:g_comp_par}. This method is also known as \emph{parametric G-computation} \cite{Robins.2009}. We use the algorithm from \cite{McGrath.2020} as another baseline. Here, the conditional densities are essentially modelled as conditional normal distributions, where both mean and variance are estimated by generalized linear regression models.

\subsubsection{Structural nested mean model (SNMM) \cite{Robins.1994}}

A structural nested mean model specifies the marginal effects of the treatment interventions at each time step. In our case, we consider the model given by
\begin{equation}\label{eq:snmm}
    \E\left[Y_{T+1}(\bar{a}_t, 0) - Y_{T+1}(\bar{a}_{t-1}, 0)\right] = \beta_t a_t
\end{equation}
for all $t \in \{1,\dots,T\}$ with parameters $\beta = (\beta_1, \dots, \beta_T)$. Here, $Y_{T+1}(\bar{a}_t, 0)$ denotes the potential outcome that is observed if the treatment intervention $\bar{a}_t$ is applied until time $t$, and no treatment is applied afterwards.

SNMM uses a method called g-estimation \cite{Robins.2009} to estimate the model parameters $\beta$. G-estimation is based on solving certain estimating equations, which is implied by Eq.~\eqref{eq:snmm} in combination with the assumptions from Sec.~\ref{sec:setting}.  For our experiments, we use the g-estimation method from \cite{Vansteelandt.2016} to obtain estimates $\hat{\beta}$.


\subsection{LTMLE}

LTMLE \cite{vanderLaan.2012} extends iterative G-computation (Algorithm~\ref{alg:iter_gcomp}) by targeting the estimates $\hat{Q}_t^a$ after each iteration step $t$. For a detailed description, we refer to \cite{vanderLaan.2012} and \cite{vanderLaan.2018}.

In our experiments, we consider two different variants of LTMLE: variant~(\lowroman{1}) estimates the conditional expectations $Q_t^a$ and the propensity scores $g_t$ with generalized linear models, and variant~(\lowroman{2}) applies the super learner algorithm \cite{vanderLaan.2007}, which uses $k$-fold cross-validation to find an optimally weighted combination of machine learning base models. We set $k=3$ and use a generalized linear model, random forest, xgboost, and a generalized additive model (GAM) with regression splines as base models.

\subsection{Deep learning for ITE estimation}

We included state-of-the-art baselines that predict future potential outcomes conditional on the patient trajectories $\mathcal{H}_t$ using deep learning. These are RMSNs \cite{Lim.2018}, CRN \cite{Bica.2020} and G-Net \cite{Li.2021}. To this end, we implemented all models as described in the respective references. Of note, these baselines aim at predicting individual counterfactual outcomes rather than estimating average causal effects. Because of that, we then obtain the ACE by averaging the respective predicted outcomes and subsequently subtracting them. For hyperparameter tuning, we refer to Appendix~\ref{sec:appendix_hyper}.

\subsubsection{RMSNs \cite{Lim.2018} and CRN \cite{Bica.2020}}

Both baselines are based on an encoder-decoder architecture, and consider the setting where treatment interventions $\bar{a}_{(t, t+ \tau -1)} = (a_t, \dots, a_{t+\tau-1})$ are applied from some time $t$ to $t + \tau - 1$. The encoder builds a representation $\Phi_t$ of the patient trajectory $\mathcal{H}_t$. This is then used by the decoder together with the treatment intervention $\bar{a}_{(t, t+ \tau -1)}$ to predict the future potential outcome $Y_{t+\tau} (\bar{a}_{(t, t+ \tau -1)})$. In the encoder, we set $\mathcal{H}_1$ = $X_1$ as input so that treatment interventions can span the complete time frame from $t = 1$ to $\tau = T$.


RMSNs address time-varying confounding by re-weighting the training loss using inverse probability of treatment weights similar to Eq.~\eqref{eq:sw}. These weights are estimated using two separate LSTMs for the nominator and denominator. In contrast, CRN adopts adversarial training techniques to build balanced representations that are non-predictive of the treatment.

\subsubsection{G-Net \cite{Li.2021}}

G-Net uses the G-formula from Eq.~\eqref{eq:g_comp_par} conditioned on the history $\mathcal{H}_1$ to estimate the outcomes $Y_{T+1} (\bar{a}_{T})$ via Monte Carlo sampling (we use $100$ samples). For this purpose, the conditional densities $f_{X_t \vert \bar{X}_{t-1}, \bar{A}_{t-1}}$ are estimated by learning the corresponding conditional expectations $\E[X_t \vert \bar{X}_{t-1}, \bar{A}_{t-1}]$ via an LSTM-based model. One can then sample from $f_{X_t \vert \bar{X}_{t-1}, \bar{A}_{t-1}}$ by drawing from the empirical distributions of the residuals on some holdout set that is not used to estimate the conditional expectations. We used 20\,\% of the training data for the holdout dataset.


\subsection{Implementation details for deep learning models}
We implemented all deep learning models (including DeepACE) using the PyTorch Lightning framework. We incorporated variational dropout \cite{Gal.2016} into the LSTMs. In particular, this allows us to provide uncertainty estimates using DeepACE. We used the Adam optimizer \cite{Kingma.2015} with 100 epochs. The feed-forward neural networks are set to one layer each, and the layer sizes are subject to hyperparameter tuning. Details on our hyperparameter tuning are in Appendix~\ref{sec:appendix_hyper}.

\section{Hyperparameter tuning details}
\label{sec:appendix_hyper}

We use Pytorch lightning in combination with the Optune package for our hyperparameter tuning implementation. We performed hyperparameter tuning for all deep learning models (including DeepACE) on all datasets by splitting the data into a training set (80\%) and a validation set (20\%). We then performed 30 random grid search iterations and chose the set of parameters that minimized the factual MSE on the validation set. The hyperparameter search ranges are shown in Table~\ref{t:hyper}.

\begin{table}[h]
\caption{Hyperparameter tuning ranges.}
\centering
\label{t:hyper}
\scriptsize
\begin{tabular}{ll}
\noalign{\smallskip} \toprule \noalign{\smallskip}
\textsc{Hyperparameter} & \textsc{Tuning range} \\
\midrule
Hidden layer size(es) & $p$, $2p$, $3p$, $4p$\\
Learning rate & $\sim \mathrm{log}\text{-}\mathrm{unif}(0.0001,0.001)$\\
Batch size &$64$, $128$ \\
Dropout probability & $0$, $0.1$, $0.2$, $0.3$\\
\bottomrule
\multicolumn{2}{l}{$p =$ network input size}
\end{tabular}
\end{table}

After training models with optimal hyperparameters, we used the full datasets to estimate the ACEs. Each model is trained over 100 epochs with the Adam optimizer~\cite{Kingma.2015}. During training, all LSTM networks use variational dropout~\cite{Gal.2016}. We set the regularization parameters to $\alpha = 0.1$ and $\beta = 0.05$. Note that it is infeasible to include $\alpha$ or $\beta$ into the hyperparameter tuning process because we only have access to the factual data (and, hence, $\alpha$ and $\beta$ may shrink to zero in order to minimize the factual loss). In doing so, we are consistent with several state-of-the-art methods for causal effect estimation \cite{Shi.2019, Bica.2020}.

\section{Datasets}
\label{sec:appendix_data}

In this section, we provide details on all datasets used to obtain our experimental results.
\subsection{Synthetic data}

We generate patient trajectories $\mathcal{D} = \left(\{x_t^{(i)}, a_t^{(i)}, y_{t+1}^{(i)}\}_{t=1}^T\right)_{i=1}^N$ from a data-generating process similar to \cite{Bica.2020b}.
The time-varying covariates $X_t \in \R^p$ follow a non-linear autoregressive process
{\footnotesize%
\begin{equation}\label{eq:data_synth_cov}
    X_t = \tanh \left( \sum_{i = 1}^h \left(\alpha_i \, X_{t-i} + \beta_i \gamma_i (2A_{t-i} -1)\, \right) + \epsilon_t^X \right)
\end{equation}}
for some lag $h \geq 1$, randomly sampled weights $\alpha_i, \beta_i \sim \mathcal{N}(1/(i+1), 0.02^2)$,  $\gamma_i \sim \mathrm{unif}\{-1, 1\}$ (discrete uniform distribution), and noise $\epsilon_t^X \sim \mathcal{N}_p(0, \mathrm{diag}_p(0.1^2))$. 
The treatments $A_t \in \{0,1\}$ are selected as $A_t = \mathbf{1}_{\{\pi_t > 0.5\}}$, where $\pi_t$ are treatment probabilities. They depend on the past observations via
{\footnotesize%
\begin{equation}
    \pi_t = \sigma\left(\tan\left(\prod_{i=1}^h \bar{X}_{t-i} \right)+ \frac{1}{p} Y_{t} + \epsilon_t^A\right),
\end{equation}}
where $\sigma$ denote the sigmoid function and $\epsilon_t^A \sim \mathcal{N}(0,0.2^2)$ is noise. Finally, the outcomes $Y_{t+1} \in \R$ are determined via
{\footnotesize%
\begin{equation}
    Y_{t+1} = \bar{X}_{t} + \sum_{i=1}^h \left( w_i \; (2 A_{t-i + 1}-1) \right)+\epsilon_{t+1}^Y
\end{equation}}
for weights $w_i = (-1)^{i+1} \frac{1}{i}$ and noise $\epsilon_{t+1}^Y \sim \mathcal{N}(0,0.1^2)$.

At the same time, we use the same process and noise to generate counterfactual data for three setups with different treatment interventions $\bar{a}_T$ and $\bar{b}_T$.
Here $\bar{b}_T$ is fixed to the zero-intervention (no treatment applied), and $\bar{a}_T$ is chosen as $(\Indicator({k \leq i\leq \ell}))_{i \in \{1,\dots, T\}}$ for $k \in \{1,3,5\}$ and $\ell \in \{10, 13, 15\}$. For our experiments in Table~\ref{t:results} and Table~\ref{t:results_mimic} we used lag $h= 5$ and $h= 8$.
The ground-truth ACE can subsequently be calculated as a Monte Carlo estimate.

\subsection{Semi-synthetic data}
We use the MIMIC-III dataset \cite{Johnson.2016} as the basis for our semi-synthetic dataset, which includes electronic health records from patients admitted to intensive care units.
We use a preprocessing pipeline \cite{Wang.2020} to extract 10 time-varying covariates $X_t \in \R^{10}$ over $T=15$ time steps.
Then, treatments $A_t \in \{0,1\}$ are simulated as binary variables with treatment probabilities
{\footnotesize%
\begin{equation}
\begin{split}
    \pi_t = & \sigma\left(\sum_{i=1}^h \left( \frac{(-1)^{i}}{1-i} \left( \bar{X}_{t-i} + \tanh(Y_{t-i}) \right) \right) \right. \\ & \left.  - \tanh\left(\ell_{t-1} - \frac{T}{2}\right) + \epsilon_t^A \right),
\end{split}
\end{equation}}
where $\epsilon_t^A \sim \mathcal{N}(0,0.5^2)$ is noise and $\ell_t$ is the current treatment level, defined by $\ell_t = \ell_{t-1} + 2(A_t - 1) \bar{X}_t \tanh(Y_t)$ and initialization $\ell_0 = T/2$.
Finally, outcomes are generated via
{\footnotesize%
\begin{equation}
\begin{split}
    Y_{t+1} = & 5 \sum_{i=1}^h\left( \frac{(-1)^{i}}{1-i} \tanh\left(\sin(\bar{X}_{t-i}A_{t-i} + \cos(\bar{X}_{t-i} A_{t-i}) \right) \right) \\ & + \epsilon_t^Y,
\end{split}
\end{equation}}
where $\epsilon_t^Y \sim \mathcal{N}(0,0.1^2)$ is noise.

\subsection{LBP data}
We analyze data from $N = 928$ patients with low back pain (LBP) \cite{Nielsen.2017}. The dataset consists of pain as time-varying outcomes recorded via assessments at $T=3$ time steps over the course of a year. Pain intensity is our outcome of interest, $Y_{t+1}$. As a treatment, we consider whether a patient has been doing physical labor or was exempt such as by recommendation of a medical professional. Covariates are given by perceived disability and 127 risk factors (\eg, age, gender). Hence, we are interested in the causal effect of whether patients have been allowed/disallowed to do physical labor on pain intensity. We estimate average causal effects for two patient cohorts, namely (1)~severe LBP and (2)~mild LBP \cite{Nielsen.2017}. We repeated this for $K=20$ evaluations with variational dropout for uncertainty estimation.

\section{Additional experimental results for static baselines}\label{sec:appendix_static}

There are several methods for causal effect estimation in the static setting. However, these do not take into account time-varying confounding and are biased in the longitudinal setting. Because of that, we refrained from reporting them in our main paper. To show that DeepACE also outperforms static methods, we implemented four state-of-the-art methods for estimating static causal effects and evaluated them on our synthetic and semi-synthetic datasets. Here, we follow \cite{Liu.2020}: we consider DragonNet \cite{Shi.2019}, TARNet \cite{Shalit.2017}, double machine learner (DML) \cite{Chernozhukov.2018}, and doubly robust learner (DR) \cite{Foster.2019}. The latter two methods are meta-learners, which we instantiate via causal forests \cite{Wager.2018}. Results are in Table~\ref{t:results_static}. Overall, DeepACE is superior by a large margin.

\begin{table*}[ht]
\caption{Results for state-of-the-art baselines for treatment effect estimation in static settings (mean $\pm$ standard deviation).}

\tiny
\label{t:results_static}
\scriptsize
\resizebox{\textwidth}{!}{\begin{tabular}{l|ccc|ccc}
\noalign{\smallskip} \toprule \noalign{\smallskip}
\textsc{Method}& \multicolumn{3}{c|}{\textsc{Synthetic data}}& \multicolumn{3}{c}{\textsc{Semi-synthetic data}} \\
\midrule
& Setup 1 & Setup 2 & Setup 3 & Setup 1 & Setup 2 & Setup 3 \\
\midrule
DragonNet \cite{Shi.2019} & $0.15 \pm 0.03$& $0.51 \pm 0.14$ & $1.17 \pm 0.49$ & $2.40 \pm 1.57$ &$3.75 \pm 2.37$ &$2.37 \pm 1.85$\\
TARNet \cite{Shalit.2017}& $0.12 \pm 0.02$& $0.46 \pm 0.17$ & $1.17 \pm 0.49$& $0.34 \pm 0.16$& $0.44 \pm 0.18$&$0.59 \pm 0.35$ \\
DML (causal forest) \cite{Chernozhukov.2018}& $0.17 \pm 0.09$ & $0.43 \pm 0.32$ & $0.66 \pm 0.77$ & $0.78 \pm 1.24$ & $0.54\pm 0.53$ &$1.06 \pm 0.08$\\
 DR (causal forest) \cite{Foster.2019}& $0.16 \pm 0.05$& $0.43 \pm 0.35$ & $0.61 \pm 0.86$& $0.55 \pm 0.47$ & $0.77 \pm 1.24$&$0.78 \pm 1.24$\\
\noalign{\smallskip} \bottomrule \noalign{\smallskip}
DeepACE (ours) & $\boldsymbol{0.04 \pm 0.02}$ & $\boldsymbol{0.09 \pm 0.07}$ & $\boldsymbol{0.18 \pm 0.08}$& $\boldsymbol{0.18 \pm 0.14}$& $\boldsymbol{0.12 \pm 0.14}$&$\boldsymbol{0.16 \pm 0.10}$\\
\bottomrule
\multicolumn{7}{l}{lower $=$ better (best in bold)}
\end{tabular}}
\vspace{-0.3cm}
\end{table*}

\end{document}